\documentclass[conference]{IEEEtran}
\IEEEoverridecommandlockouts
\usepackage{cite}
\usepackage{amsmath,amssymb,amsfonts}
\usepackage{algorithmic}
\usepackage{algorithm}
\usepackage{graphicx}
\usepackage{textcomp}
\usepackage{xcolor}
\usepackage{graphicx}
\usepackage{pdfpages}
\usepackage{url}
\usepackage{hyperref}

\newlength\myindent
\def\BibTeX{{\rm B\kern-.05em{\sc i\kern-.025em b}\kern-.08em
    T\kern-.1667em\lower.7ex\hbox{E}\kern-.125emX}}
\begin{document}

\title{Personalized Federated Learning via Heterogeneous Modular Networks}

\author{
\IEEEauthorblockN{Tianchun Wang$^1$, Wei Cheng$^2$, Dongsheng Luo$^3$, Wenchao Yu$^2$, Jingchao Ni$^4$, Liang Tong$^2$,\\ Haifeng Chen$^2$, Xiang Zhang$^1$}
\IEEEauthorblockA{\textit{$^1$The Pennsylvania State University},
\textit{$^2$NEC Laboratories America}, \textit{$^3$Florida International University}, \textit{$^4$AWS AI Labs, Amazon}}
\{tkw5356, xzz89\}@psu.edu, \{weicheng, wyu, ltong, haifeng\}@nec-labs.com, dluo@fiu.edu, jingchni@amazon.com
}

\maketitle

\begin{abstract}
Personalized Federated Learning (PFL) which collaboratively trains a federated model while considering local clients under privacy constraints has attracted much attention. Despite its popularity, it has been observed that existing PFL approaches result in sub-optimal solutions when the joint distribution among local clients diverges. To address this issue, we present Federated Modular Network (FedMN), a novel PFL approach that adaptively selects sub-modules from a module pool to assemble heterogeneous neural architectures for different clients. FedMN adopts a light-weighted routing hypernetwork to model the joint distribution on each client and produce the personalized selection of the module blocks for each client. To reduce the communication burden in existing FL, we develop an efficient way to interact between the clients and the server. We conduct extensive experiments on the real-world test beds and the results show both effectiveness and efficiency of the proposed FedMN over the baselines.
\end{abstract}


\section{Introduction}
Federated Learning (FL) emerges
as a prospective solution that facilitates distributed collaborative learning without disclosing original training data whilst
naturally complying with the government regulations \cite{lim2020federated,aledhari2020federated}. 
In practice, the problem of data heterogeneity deteriorates the performance of the global FL model on individual
clients due to the lack of solution personalization. 
To tackle with it, researchers focus on the Personalized Federated Learning (PFL) which aims to make the global model fit the distributions on most of the devices \cite{wang2019federated,t2020personalized}.
The vanilla PFL approaches first learn a global model and then locally adapt it to each client by fine-tuning the global parameters \cite{sim2019investigation,mansour2020three}. In this case, the trained global model can be regarded as a meta-model ready for further personalization of each local client. In order to build a better meta-model, many efforts have been done to bridge the FL and the Model Agnostic Meta Learning (MAML) \cite{acar2021debiasing,khodak2019adaptive,fallah2020personalized}. However, the global generalization error typically does not decrease much \cite{marfoq2021federated} for these approaches. Thus, the performance can not be significantly improved. Another line of research focuses on jointly training a global model and a local model for each client to achieve personalization \cite{corinzia2019variational,deng2020adaptive}. This strategy does not perform well on the clients whose local distributions are far from their average. Cluster-based PFL approaches \cite{ghosh2020efficient} address this issue by grouping the clients into several clusters. The clients in a cluster share the same model while those belonging to different clusters have different models. Unfortunately, the model trained in one cluster will not benefit from the knowledge of the clients in other clusters, which limits the capability to share knowledge and therefore results in a sub-optimal solution. 

An alternative strategy is adopting the Multi-Task Learning (MTL) framework to train a PFL model \cite{smith2017federated,t2020personalized}. However, most existing efforts did not consider the difference in conditional distribution between clients. It is an important problem when building a federated model. For example, labels sometimes reflect sentiment. Some users may label a laptop as cheap while others label it as expensive. This \emph{conditional distribution heterogeneity} problem will cause the model inaccurate on some clients where the $p(\mathbf y|\mathbf x)$ is far from the average. To address the problem, a recent work \cite{marfoq2021federated} assumes the data distribution of each client is a mixture of $M$ underlying distributions and proposes a flexible framework in which each client learns a combination of $M$ shared components with different weights. It optimizes the varying conditional distribution $p_i(\mathbf y|\mathbf x)$ under the assumption that the marginal distribution $p_i(\mathbf x)=p(\mathbf x)$ is the same for all clients (Assumption 2 in \cite{marfoq2021federated}). This assumption, however, is restricted. For instance, in handwriting recognition, users who write the same words might still have different stroke widths, slants, etc. In this cases, $p_i(\mathbf x) \neq p_j(\mathbf x)$ for client $i$ and $j$. Other works \cite{shlezinger2020communication,sattler2019robust} either assume the marginal distribution $p_i(\mathbf x)$ or the conditional distribution $p_i(\mathbf y|\mathbf x)$ the same across clients. In reality, data on each client may be deviated from being identically distributed, say, $P_i \neq P_j$ for client $i$ and $j$. That is, the joint distribution $P_i(\mathbf x,\mathbf y)$ (can be 
rewritten as $P_i(\mathbf y|\mathbf x)P_i(\mathbf x)$ or $P_i(\mathbf x|\mathbf y)P_i(\mathbf y)$) may be different across clients. We call it the ``\textbf{joint distribution heterogeneity}" problem. Existing approaches \cite{shlezinger2020communication,sattler2019robust} fail to completely model the difference of joint distribution between clients because they assume one term to be the same while varying the other one. Moreover, to accommodate different data distributions, the homogeneous model would be too large so that the given prediction power can be satisfied. Thus, the communication costs between the server and clients would be huge. In this case, 
communication would be a key bottleneck to consider
when developing FL methods. To this end, it is desirable to design an effective PFL model to accommodate heterogeneous clients in an efficient.

To solve the aforementioned problems, in this paper, we propose a novel \underline{F}ederated \underline{M}odular \underline{N}etworks (FedMN) approach, which personalizes heterogeneous clients efficiently. The main idea is that we implicitly partition the clients by modeling their joint distribution into clusters and the clients in the same cluster have the same architecture. Specifically, a shared module pool with layers of module blocks (e.g., MLPs or ConvNets) is maintained in the server. Each client 
decides in each update to assemble a personalized model by selecting a combination of the blocks from the module pool. We adopt a light-weighted routing hypernetwork with differentiable routers to generate the decision of module block selection for each client. The routing hypernetwork considers the joint distribution $p_i(\mathbf x,\mathbf y)$ for client $i$ by taking the joint distribution of the data set as the input. A decision parameterized by the routing hypernetwork is a vector of discrete variables following the Bernoulli distribution. It selects a subset of the blocks from the module pool to form an architecture for each client. Clients with similar decisions will be implicitly assigned to the same cluster in each communication round. 
The proposed FedMN enables a client to upload only a subset of model parameters to the server, which decreases the communication burden compared to traditional FL algorithms. To sum up, our contributions are as follows: 1) We address the problem of joint distribution heterogeneity in the personalized FL and propose a FedMN approach to alleviate this issue. 2) We develop an efficient way to selectively upload model parameters which decreases the communication cost between clients and the server. 3) Extensive experiments on the real-world datasets show both the effectiveness and the efficiency of our proposed FedMN compared to the state-of-the-arts.

\section{Methodology}
Our method adopts modular networks which consist of a group of encoders in the first layer and multiple modular blocks in the following layers. The connection decisions between blocks are made by a routing hypernetwork.
\subsection{Modular Networks}
\label{sec:modular}
The modular networks first encode the data feature into low-dimensional embeddings by a group of encoders motivated by ~\cite{kirsch2018modular,pahuja2019structure,zhmoginov2021compositional}. 
Then, personalized feature embeddings are obtained by discovering and assembling a set of modular blocks in different ways for different clients. 
The modular networks has $L$ layers and the $l$-th layer has $n_l$ blocks of sub-network. The encoders in the 1st layer are $n_1$ independent blocks which learns feature embeddings for each client. Formally, let $\mathbf{x}_i$ be the $i$-th sample, we get the feature embedding $\mathbf{z}_i^{(j)}$ after the $j$-th encoder is applied
\begin{small}
\begin{align}
    \label{eq:encoder}
    \setlength{\abovedisplayskip}{2pt}
\setlength{\belowdisplayskip}{2pt}
\mathbf{z}^{(j)}_i = Encoder^{(j)}\left(\mathbf{x}_i \right), \quad j=1,...,n_1.
\end{align}
\end{small}
The choices of encoder networks are flexible. For example, one can adopt CNNs as encoders for image data and transformers for text data. 

The set of feature embeddings $\{\mathbf{z}_i^{(1)},...,\mathbf{z}_i^{(n_1)}\}$ of data point $\mathbf{x}_i$ resulting from the encoders in the 1st layer is the input of the following modular sub-networks constructed by a subset of the modular blocks. There are $L-1$ layers of blocks in the sub-networks and each one is independent of the others. 
Each modular block $j$ in layer $l$ receives a list of $n_{l-1}$ tensors of feature embeddings from the modular sub-networks in the layer $l-1$. We use MLPs as the modular blocks in this paper and each pair of them in successive layers may be connected or not.  At most, there are $\mathcal{E}$ possible connection paths between modular blocks and $\mathcal{E}=\sum_{j=1}^{L-1}n_j n_{j+1}+n_L.$ To determine which path would be connected, we need to learn a decision $\mathbf{V}_m \in \mathbb{Z}_2^{\mathcal{E}}$ for client $m$. Each element $v_i^{(m)} \in \mathbf{V}_m$ is a binary variable with values chosen from $\{0,1\}$.  $v_i^{(m)}=1$ indicates that the path between two blocks is connected, and 0 otherwise. Since some blocks may not have connected paths, $\mathbf{V}_m$ also determines which subset of blocks will be selected from the modular pool for each client. Therefore, after obtaining $\mathbf{V}_m$, the architecture for a client is determined. \begin{figure}[t]
\vspace{-0.8in}
\centering
\includegraphics[width=0.75\textwidth, trim = 0cm 5cm 7cm 0cm]{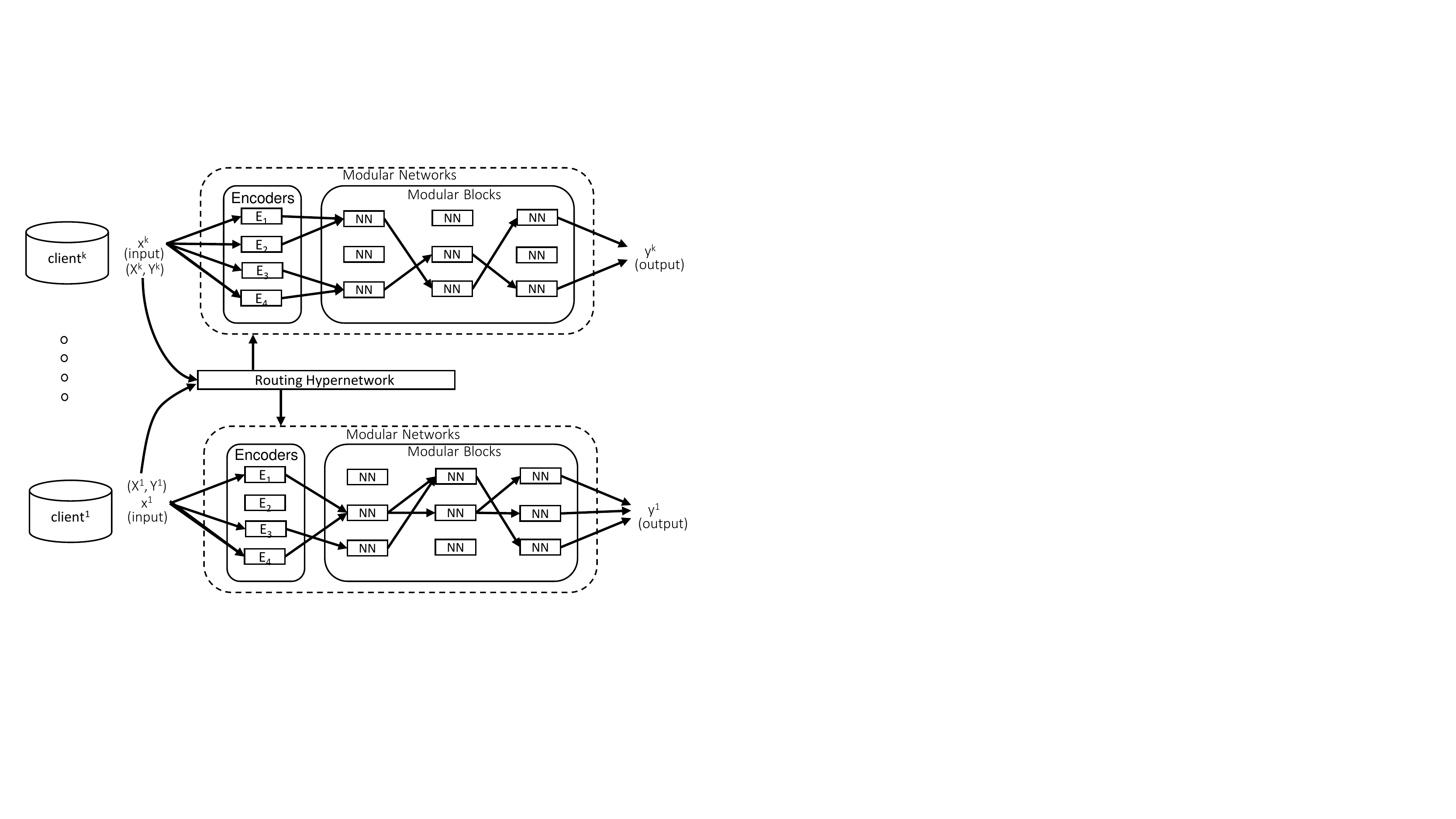}
\caption{The FedMN architecture. The modular networks consist of a group of encoders in the first layer and modular blocks in the following layers. The connection paths between blocks are determined by the routing hypernetwork. The input of modular networks is in a sample-wise way, while the input of routing hypernetwork is the full local dataset for each client.}
\vspace{-0.1in}
\label{fig:modularnets}
\end{figure}

\vspace{-0.02in}
\subsection{The Learning Objective}
\vspace{-0.02in}
With the defined modular networks, we can formally define our learning objective. Suppose there are $M$ clients where each client has local dataset $\mathcal{D}_m=\{(\mathbf{x}_i,y_i)\}_{i=1}^{\lvert \mathcal{D}_m\rvert}$.
In the FedMN framework, after getting $\mathbf{V}_m$, the architecture of the modular network for client $m$ is fixed at an epoch during local updating. 
We let $f_{\theta}$ is the model parameterized by $\theta$ that contains the parameters of both modular networks and the routing hypernetwork. When making a prediction, we have $\hat{y}_i=f_{\theta}\left(\mathbf{x}_i; \mathbf{V}_m \right)$. Then, the empirical
risk of FedMN is
\begin{small}
\begin{align}
    \label{eq:fedmn_obj}
    \setlength{\abovedisplayskip}{2pt}
\setlength{\belowdisplayskip}{2pt}
& \min_{\theta,\{\mathbf{V}_m\}_{m=1}^M} \sum_{m=1}^M \frac{\lvert\mathcal{D}_m\rvert}{\lvert\mathcal{D}\rvert}\mathcal{L}_m\left( \theta,\mathbf{V}_m\right),\\ 
& \text{where} \quad \mathcal{L}_m\left(\theta,\mathbf{V}_m \right)=\frac{1}{\lvert\mathcal{D}_m\rvert}\sum_{(\mathbf{x}_i,y_i) \in \mathcal{D}_m} \ell\left(
f_{\theta}\left(\mathbf{x}_i; \mathbf{V}_m  \right)
,y_i \right). \nonumber
\end{align}
\end{small}

However, the direct optimization of the objective in (\ref{eq:fedmn_obj}) is intractable as there are $2^{\mathcal{E}}$ candidates for each $\mathbf{V}_m$. Thus, we consider a relaxation by assuming that the decision of each connection path in  $v_i^{(m)} \in \mathbf{V}_m$ is conditionally independent to each other. Formally, we have
\begin{small}
\begin{align}
    \label{eq:factorV}
    \setlength{\abovedisplayskip}{2pt}
\setlength{\belowdisplayskip}{2pt}
P\left(\mathbf{V}_m \right)=\prod_{v_i^{(m)} \in \mathcal{E}} P\left(v_i^{(m)} \right).
 \end{align}
\end{small}

A straightforward instantiation of $P\left(v_i^{(m)} \right)$ is the Bernoulli distribution $v_i^{(m)} \sim Bern\left(\pi_i^{(m)} \right)$. $P\left(v_i^{(m)}=1 \right)=\pi_i^{(m)}$ is the probability that the $i$-th path exists in $\mathbf{V}_m$. With this relaxation, we thus can rewrite the
objective in (\ref{eq:fedmn_obj}) as
\begin{small}
\begin{align}
    \label{eq:exp_obj}
    \setlength{\abovedisplayskip}{2pt}
\setlength{\belowdisplayskip}{2pt}
& \min_{\theta,\{\mathbf{V}_m\}_{m=1}^M} \sum_{m=1}^M \frac{\lvert\mathcal{D}_m\rvert}{\lvert\mathcal{D}\rvert}\mathcal{L}_m\left( \theta,\mathbf{V}_m\right)\\ \nonumber
& \approx \min_{\mathbf{\theta},\{\mathbf{\Pi}_m\}_{m=1}^M} \sum_{m=1}^M \frac{\lvert\mathcal{D}_m\rvert}{\lvert\mathcal{D}\rvert} \mathbb{E}_{\mathbf{V}_m \sim q\left(\mathbf{\Pi}_m \right)}\left[\mathcal{L}_m\left(\mathbf{\theta}, \mathbf{\mathbf{V}}_m \right)\right], 
\end{align}
\end{small}
\noindent where $q\left(\mathbf{\Pi}_m\right)$ is the distribution of the decision variable parameterized by $\pi^{(m)}$'s.

\subsection{Efficient Computation with Reparameterization}

Due to the binary nature of $\mathbf{V}_m$, it is impractically to optimize (\ref{eq:exp_obj}) with gradient-based back prorogation. To enable efficient computation, we further approximate the binary vector $\mathbf{V}_m \in \mathbb{Z}_2^{\mathcal{E}}$ with a continuous real-valued vector in $[0, 1]^\mathcal{E}$. In practice, we approximate each Bernoulli distribution $v_i^{(m)} \sim Bern\left(\pi_i^{(m)} \right)$ with a  binary concrete distribution~\cite{jang2016categorical,maddison2016concrete}. Formally, letting $\sigma(\cdot)$ as the Sigmoid function, we have
\begin{small}
\begin{equation}
\begin{aligned}
\label{eq:concrete}
    & v_i^{(m)} \approx \sigma((\log \epsilon - \log (1-\epsilon) + \log\frac{\pi^{(m)}_i}{1-\pi_i^{(m)}})/\tau), \\
    & \text{where} \quad \epsilon \sim \text{Uniform}(0,1).
\end{aligned}
\end{equation}
\end{small}
The hyper-parameter $\tau$ is a temperature variable to trade-off between approximation and binary output. When the temperature $\tau$ approaches to 0, the binary concrete distribution of $v_i^{(m)}$ in (\ref{eq:concrete}) converge to the Bernoulli distribution $v_i^{(m)} \sim Bern\big(\pi_i^{(m)} \big)$.  With reparameterization, combining (\ref{eq:exp_obj}) and (\ref{eq:concrete}) we have the learning objective
\begin{small}
\begin{align}
    \label{eq:continuous_obj}
    \setlength{\abovedisplayskip}{2pt}
\setlength{\belowdisplayskip}{2pt}
\min_{\mathbf{\theta},\{\mathbf{\Pi}_m\}_{m=1}^M} \sum_{m=1}^M \frac{\lvert\mathcal{D}_m\rvert}{\lvert\mathcal{D}\rvert} \mathbb{E}_{\epsilon \sim \text{Uniform}(0,1)}\left[\mathcal{L}_m\left(\mathbf{\theta}, \mathbf{\mathbf{V}}_m \right)\right],
\end{align}
\end{small}
When temperature $\tau>0$, the objective function in (\ref{eq:continuous_obj}) has a well-defined gradient that enables efficient optimization with back-propagation.  
\subsection{Routing Hypernetwork}
\label{sec:routing}
In this part, we introduce the routing hypernetwork that automatically learns $\mathbf{\Pi}_m$ from the joint distribution. The joint embedding of $\mathbf{X}, \mathbf{Y}$ is given by
\begin{small}
\begin{align}
    \label{eq:jointemb_two}
    \setlength{\abovedisplayskip}{2pt}
\setlength{\belowdisplayskip}{2pt}
\hat{\mathcal{C}}_{\mathbf{X}\mathbf{Y}}=\frac{1}{n^m}\sum_{i=1}^{n^m}  \phi^x\left(\mathbf{x}_i \right)\phi^y\left(y_i \right),
\end{align}
\end{small}
where $n^m$ is the number of samples in client $m$.
As mentioned in \cite{song2013kernel}, the mappings $\phi^x(\mathbf{x})$ and $\phi^y(y)$ are flexible. In this paper, we denote $\theta_h$ as the parameters used in the routing hypernetwork, which is a part of the model parameters $\theta$. We parameterize the feature mappings by neural networks similar to \cite{DeepDist} thus (\ref{eq:jointemb_two}) results in
\begin{small}
\begin{align}
    \label{eq:jointemb_two_par}
    \setlength{\abovedisplayskip}{2pt}
\setlength{\belowdisplayskip}{2pt}
\hat{\mathcal{C}}_{\mathbf{X}\mathbf{Y},\theta_h}=\frac{1}{n^m}\sum_{i=1}^{n^m}  \phi_{\theta_h}^x\left(\mathbf{x}_i \right)\phi_{\theta_h}^y\left(y_i \right).
\end{align}
\end{small}
Then, two fixed-size vector representation of a dataset is thus provided
by the averaged output of the two neural networks $\phi^x_{\theta_h}: \mathbf{x} \mapsto \mathbb{R}^{d_x}$ and $\phi^y_{\theta_h}: \mathbf{y} \mapsto \mathbb{R}^{d_y}$. By the Universal Approximation Theorem (Chapter 6 in \cite{Goodfellow-et-al-2016}), we concatenate $\phi_{\theta_h}^x\left(\mathbf{x}_i\right)$ and $\phi_{\theta_h}^y\left(y_i \right)$ and adopt a single-layer perceptron $h_{\theta_h}: \mathbb{R}^{d_x+d_y} \mapsto \mathbb{R}^{\mathcal{E}}$, where $\mathcal{E}$ is the total possible number of paths between successive modules, thus the product operator in (\ref{eq:jointemb_two_par}) can be approximated 
by 
\begin{small}
\begin{align}
    \label{eq:jointemb_two_par_p}
    \setlength{\abovedisplayskip}{2pt}
\setlength{\belowdisplayskip}{2pt}
\hat{\mathcal{G}}_{\theta_h}\left(\mathbf{X},\mathbf{Y} \right)=\frac{1}{n^m}\sum_{i=1}^{n^m}  h_{\theta_h}\left(\left[\phi_{\theta_h}^x\left(\mathbf{x}_i \right), \phi_{\theta_h}^y\left(y_i \right)\right]\right),
\end{align}
\end{small}
which results in a vector of joint embedding of the local dataset at client $m$. Then, $\mathbf{\Pi}_m$ can be obtained by
\begin{small}
\begin{align}
    \label{eq:jointemb_pi}
    \setlength{\abovedisplayskip}{2pt}
\setlength{\belowdisplayskip}{2pt}
\mathbf{\Pi}_m &=
\sigma\left(\hat{\mathcal{G}}_{\theta_h}\left(\mathbf{X},\mathbf{Y} \right)\right)\\ \nonumber
&=\sigma\left(\frac{1}{n^m}\sum_{i=1}^{n^m}  h_{\theta_h}\left(\left[\phi_{\theta_h}^x\left(\mathbf{x}_i \right), \phi_{\theta_h}^y\left(y_i \right)\right]\right)\right).
\end{align}
\end{small}
\subsection{Communication Strategy}
\label{sec:commstra}
Since $\mathbf{V}_m$ determines the connection paths between blocks, some blocks may not have connections with other ones. To clarify the massage passing between blocks, we denote the connection paths between blocks in the layer $(l-1)$ and the layer $l$ as $\mathbf{C}_m\in \mathbb{Z}_2^{n_l \times n_{l-1}}$, with the element $C_{jk}^{(m)} \in \{0, 1\}$ in its j-th row and k-th column. Letting $\mathbf{u}_j^{(l)}$ be the input tensor for the $j$-th block in the layer $l$, and $\mathbf{\tilde{u}}_j^{(l)}$ be its output, we have
\begin{small}
\begin{align}
    \label{eq:block_input}
    \setlength{\abovedisplayskip}{2pt}
\setlength{\belowdisplayskip}{2pt}
\mathbf{u}_j^{(l)}=\left\{
\begin{array}{cl}
\frac{\sum_{k=1}^{n_{l-1}} C_{jk}^{(m)}\mathbf{\tilde{u}}_k^{(l-1)}}{\sum_{k=1}^{n_{l-1}}C_{jk}^{(m)}},   &{\text{if}\quad \sum_{k=1}^{n_{l-1}} C_{jk}^{(m)}\neq 0,}\\
 0,    &{\text{otherwise.}}
\end{array} \right.
\end{align}
\end{small}
To decrease the number of model parameters transmitted between clients and the server, we develop a block-wise strategy for clients to upload the local models to the server and copy them from the server. In detail, when we have the decision $\mathbf{V}_m$, we know that the inputs for some blocks are 0's from (\ref{eq:block_input}). Therefore, some blocks are still active whose input is not all 0's.
In total, we denote there are $B$ blocks in the modular network, where $B=n_2+...+n_L$.
Let  $\mathbf{a}_m \in\mathbb{Z}_2^B$ to denote which blocks are active for the local model at client $m$, with the element $a_i^{(m)}=1$ if the input for the $i$-th block is not 0 while $a_i^{(m)}=0$ otherwise. When uploading the model to the server, the client only uploads the active blocks whose  $a_i^{(m)}=1$. When copying the model from the server, the client only copies the parameters of active blocks from the global model. This strategy will significantly reduce unnecessary communication costs between clients and the server.
In each round, the local model at a client is uploaded to the server. After a round, all the clients upload their local models to the server for aggregation. For the modular networks, the aggregation however is in a block-wise manner. Specifically, let $\theta_i^{(m)}$ be the model parameters of the $i$-th modular block for client $m$, and the server performs the aggregation to obtain the global parameter of the $i$-th modular block by $\theta_i=\sum_{m=1}^M \frac{\lvert \mathcal{D}_m \rvert}{\lvert \mathcal{D} \rvert} a_i^{(m)} \theta_i^{(m)}$. In detail, we list the federated learning process of FedMN in Algorithm 1.

\begin{algorithm}

\label{alg:fedmnalg}
\begin{small}
    \centering
    \caption{FedMN: Federated Modular Networks Algorithm}
    \begin{algorithmic}[1]
        \STATE {\bfseries Input:}  Number of clients $M$; local dataset $\{\mathcal{D}_m\}_{m=1}^M$, where $\mathcal{D}_m=\{\left(\mathbf{x}_i, y_i\right)\}_{i=1}^{\lvert\mathcal{D}_m\rvert}$; number of layers of modular network $L$; number of modular blocks in each layer $\{n_l\}_{l=1}^L$; number of communication rounds $T$; number of local epochs $K$; learning rate $\eta$.
        \STATE {\bfseries Output:} Local models $\theta_{m}^T$ for $m\in[M]$; global model $\theta^T$; local decisions $\mathbf{V}_m^T$ for $m\in[M]$.
        \STATE Server initializes the global modular pool $\{n_l\}_{l=1}^L$. 
        \FOR{\textbf{iterations} $t=1,...,T$}
            \FOR{\textbf{clients} $m=1,...,M$}
                \STATE Get $\mathbf{\Pi}_m$ with local dataset $\mathcal{D}_m$ as in (\ref{eq:jointemb_pi});
                \STATE $\mathbf{V}_m \gets \text{binaryConcrete}\left(\mathbf{\Pi}_m \right)$  as in (\ref{eq:concrete});
                \STATE Determine the local model $f_{\theta}$ as in Section~\ref{sec:modular};
                \STATE Client $m$ copy $\theta_{m-1}^{t-1}$ from the server as in Section~\ref{sec:commstra};
                \FOR{\textbf{local epochs} $k=1,...,K$}
                    
                        \STATE $\theta^{t-1}_{k,m} \gets \text{LocalSolver}\left(\theta^{t-1}_{k-1,m-1}, \mathbf{V}_m, \eta \right)$;
                \ENDFOR
                \STATE Client $m$ sends $\theta_{K,m}^{t-1}$ to the server as in Section~\ref{sec:commstra};
                
            \ENDFOR
            \STATE Server averages $\{\theta_{K,m}^{t-1}\}_{m=1}^M$ to get the global parameters $\theta^{t}$ as in Section \ref{sec:commstra};
            \STATE Compute the global loss as in (\ref{eq:continuous_obj});
    \ENDFOR 
    
    \textbf{Function} {LocalSolver}({$\theta^{t-1}_{k-1}, \mathbf{V}_m, \eta$}):
    \begin{ALC@g}
     \FOR{\textbf{each batch}}
            \STATE $\theta \gets \theta - \eta \nabla_{\theta}\mathbb{E}_{\mathbf{V}_m \sim q\left(\mathbf{\Pi}_m \right)}\left[\mathcal{L}_m\left(\mathbf{\theta}^{(t-1)}_k, \mathbf{\mathbf{V}}_m \right)\right]$;
        \ENDFOR
    \STATE \textbf{return} $\theta$ 
    \end{ALC@g}
    \end{algorithmic}
\end{small}
\end{algorithm}

\section{Related Work}
The standard FL approaches train a unique model for all clients~\cite{Mitigating,konevcny2016federated}. However, in the real settings, there always exits statistical heterogeneity across clients~\cite{kairouz2021advances}. Indeed, data on different clients may be collected in different ways and they may behave in non-IID and unbalanced manner~\cite{zhao2018federated,sattler2019robust,li2021federated}. Some efforts extend the FL methods for heterogeneous clients to achieve personalization~\cite{t2020personalized,tan2022towards}. Several works train a global model as a meta-model and then fine-tune the parameters for each client~\cite{sim2019investigation,jiang2019improving}, which still have difficulty for generalization~\cite{caldas2018leaf}. Clustered FL~\cite{sattler2020clustered} methods are more related to our work. These efforts assume that clients can
be assigned to several clusters and those in the same cluster share the same model. In this way, clients belonging to different clusters do not have relevant knowledge from each other, which will make the federated model sub-optimal. Different from them, our method enables all clients to have relevant knowledge shared with each other via the shared modular pool and routing hypernetwork. Another group of methods related to ours is using multi-task learning to learn personalized
models in the FL setting which allows for more nuanced relations among clients’ models~\cite{smith2017federated,vanhaesebrouck2017decentralized,caldas2018federated}. However, they did not consider the heterogeneous statistical diversity. The work in~\cite{marfoq2021federated} considers conditional distribution between clients but assumes their marginal distributions are fixed. Different from that, our approach models the heterogeneity of joint distributions across clients. 
\begin{table*}[htbp]
\vspace{-0.2in}
\centering
  \caption{Datasets and models.}\label{tab:result1}
  \vspace{-3mm}
  \begin{center}
\scalebox{0.9}{
  \begin{tabular}{|c| c| c| c |c | c |}
  \hline
\textbf{Dataset} & \textbf{Task} & \textbf{Number of clients} & \textbf{Number of samples} & \textbf{Model} & \textbf{FedMN Architecture}\\\hline
CIFAR10 & Image classification& 80 & 60,000 & MobileNet-v2 & [$2\times2\times2$]\\\hline
CIFAR100 & Image classifcation & 100 & 60,000 & MobileNet-v2 & [$2\times 2\times 2$]\\\hline
EMNIST & Handwritten character recognition & 100 & 81,425 &2-layer CNN + 2-layer FFN & [$3\times3\times3$] \\\hline
FEMNIST& Handwritten character recognition & 539& 120,772&2-layer CNN + 2-layer FFN&[$3\times3\times3$] \\\hline
\end{tabular}
}
\end{center}
\vspace{-1mm}
\end{table*}
\begin{table*}[t]
\centering
  \caption{Average test accuracy (\%) across clients.}\label{tab:result2}
  \vspace{-3mm}
  \begin{center}
\scalebox{1.1}{
  \begin{tabular}{|c| c| c| c |c |c |c |c ||c|}
  \hline
Dataset & Local & FedAvg & FedProx & FedAvg+ & ClusteredFL & pFedMe & FedEM & FedMN(Ours)\\\hline
CIFAR10 & 70.2& 78.2& 78.0&82.3&  78.6& 81.7& 83.2 & \textbf{83.4}\\\hline
CIFAR100 & 31.5 & 40.9&41.0& 39.0 &41.5& 41.8&43.6& \textbf{45.6}\\\hline
EMNIST &71.9& 82.6&83.0&83.1&82.7& 83.3&83.5&\textbf{83.8}\\\hline
FEMNIST&71.0& 78.6& 78.9&75.3& 73.5&74.9&79.9&\textbf{82.2}\\\hline
\end{tabular}
}
\end{center}
\vspace{-1mm}
\end{table*}
Modular networks are composed of modules
and each module is a function with its parameters~\cite{kirsch2018modular}. Learning an efficient set of such modules is akin to learning
a set of functional primitives, which are evident in the
natural world and can be combined to solve a given task~\cite{meyerson2017beyond}. Some recent progress has been
made on the generalization of learning across diverse domains~\cite{zhao2018modular}. However, these methods are not designed
for PFL. In addition, modular networks are clearly
related to routing networks~\cite{bayat2021coded}, where each sample selectively activates only parts of the entire network. Some routing networks based on conditional computation have been proposed to improve
computational efficiency~\cite{bengio2015conditional}. Several approaches routed the examples
through a dynamic network for multiple tasks~\cite{rosenbaum2017routing}. Unlike these methods, we consider the FL setting where data owners preserve their privacy. 

\section{Experiments}
\vspace{-0.05in}
\subsection{Datasets and Baselines}
 To evaluate our method, we conduct extensive experiments on four federated benchmark datasets: image classification on CIFAR10 and CIFAR100~\cite{krizhevsky2009learning}, handwriting character recognition on EMNIST \cite{cohen2017emnist} and FEMNIST \cite{caldas2018leaf}. We preprocessed all the datasets in the same way as in~\cite{marfoq2021federated} to build the testbed. Detailed dataset partitioning can be found in~\cite{marfoq2021federated}.
 The dataset statistics and FedMN architectures are summarized in Table~\ref{tab:result1}.
 
To show the effectiveness of our approach, we compare the proposed FedMN with the following baselines: 1) Local: a personalized model trained only
on the local dataset at each client; 2) FedAvg~\cite{mcmahan2017communication}: a generic FL method that trains a unique global model for all clients; 3) FedProx~\cite{li2020federated}: a re-parametrization of FedAvg to tackle with statistical heterogeneity in federated networks; 4) FedAvg+~\cite{jiang2019improving}: a variant of FedAvg with two stages of training and local tuning; 5) Clustered FL~\cite{sattler2020clustered}: a framework exploiting geometric properties of the FL loss surface which groups the client population into clusters using conditional distributions; 6) pFedMe~\cite{t2020personalized}: a  bi-level optimization framework designed for PFL which decouples the optimization from learning the global model; 7) FedEM~\cite{marfoq2021federated}: a federated multi-task learning approach based on the assumption that
local data distributions are mixtures of underlying distributions.

\vspace{-0.05in}
\subsection{Implementation Details}
For a fair comparison, the proposed FedMN~\footnote{\url{https://github.com/xztcwang/FedMN}} adopts the same architectures as the baseline methods for its encoder blocks. Specifically, FedMN uses a group of MobileNet-v as the encoders on CIFAR10 and CIFAR 100, 2-layer CNN as encoders on EMNIST and FEMNIST. The encoders are followed by several layers of fully connected layers with dropout as modular blocks. We show the architectures of modular networks on all datasets for the performance comparison task in Table~\ref{tab:result1}. For example, on CIFAR10 dataset the FedMN Architecture is $\left[2\times 2\times 2 \right]$.  It means that in FedMN there are 2 MobileNet-v encoders in Layer-1, followed by 2 blocks of a fully connected layer with dropout in Layer-2, and 2 linear layers in Layer-3 for output. In the proposed FedMN model, the routing hypernetwork adopts the same encoder architecture as in the modular network to embed the data feature. The training labels are encoded in one-hot and embedded by a fully connected layer. After concatenating the feature embedding and label embedding, we normalize it to input it into a fully connected layer. Then, as discussed in Sec. \ref{sec:routing}, we calculate the mean of the output embedding and perform the reparameterization trick to obtain the routing decisions. In our experiments, the total communication rounds are set to 150 and local epoch is set to 1. We choose SGD as the local solver with the learning rate tuned via grid search in the range $\{1\cdot 10^{-3}, 2\cdot10^{-3}, 5\cdot10^{-3}, 1\cdot10^{-2}, 2\cdot10^{-2}, 5\cdot10^{-2}\}$ to obtain the best performances. We use FedAvg to pretrain the encoder model in our experiments. We follow the practice in~\cite{jang2017categorical} to adopt the exponential decay strategy by starting the training with a high temperature 1.0, and anneal to a small value 0.1. All experiments are implemented using  PyTorch and TorchVision. We run all the experiments on a Linux machine with 8 NVIDIA A100 GPU, each one with 80GB memory.  
\vspace{-0.05in}
\subsection{Effectiveness Analysis}

We compare all the methods on the testbed where the datasets are processed and clients are partitioned the same as in~\cite{marfoq2021federated}. Therefore, we reuse their reported accuracy of Local, FedAvg, FedProx, FedAvg+, ClusteredFL and pFedMe. For fair comparison, the number of encoders in FedMN are set as the same as the number of components in FedEM on each dataset. Table~\ref{tab:result2} list the average classification accuracy of all methods on all datasets. The performance of our method is evaluated on the local test data on each client and we report the averaged accuracy of all clients. From the results, we observe that the personalized method FedAvg+ outperforms FedAvg and FedProx on CIFAR10 but is not as good as them on other datasets because only locally tuning the global model cannot generalize well. Clustered FL cannot outperform FedAvg+ on all datasets because without knowledge sharing between clusters, optimal personalized solutions cannot be obtained on all clients. FedEM outperforms the other baselines because it takes advantage of the heterogeneity of conditional distributions. From the table, it is obvious that our proposed FedMN outperforms all the baselines. It is because FedMN builds personalized models according to the joint data distribution of each client, which captures the heterogeneity of each client.

\subsection{Communication Efficiency}
\begin{figure}[htbp] 
\vspace{-0.2in}
 \center{\includegraphics[width=5cm]  {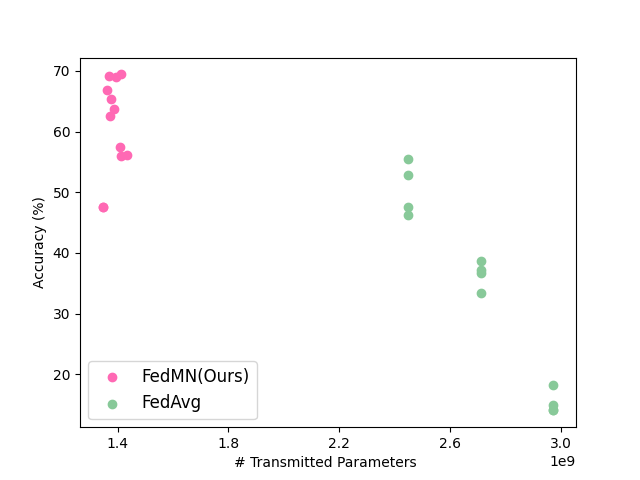}} 
 \caption{\label{fig:trans_param} Average test accuracy across clients \emph{w.r.t.} number of transmitted parameters.} \vspace{-0.1in}
 \end{figure}
To show the communication efficiency of the proposed FedMN, we compare it with the representative baseline FedAvg because we use a similar way to average model blocks for model aggregation. Note that many other baselines are using a similar strategy. For example, FedEM also needs to conduct the average on the $M$ model components. Thus, its communication costs are even larger than FedAvg. For a fair comparison, we use MLPs for both FedMN and FedAvg with the same parameter scale. In detail, we build 8-Layer, 9-Layer, and 10-Layer MLPs respectively for FedAvg. In FedMN, the number of modular blocks is set to 8, 9, and 10 respectively with different architectures and we use only one fully connected layer as the encoder for all of them. For example, for 8 modular blocks, we set the FedMN architecture as $\left[1\times 4\times 3\right]$, $\left[1\times 3\times 4\right]$, $\left[1\times 2\times 5\right]$, and $\left[1\times 5\times 2\right]$, respectively. We use a fully connected layer as the encoder in their routing hypernetworks. The number of hidden units of the MLPs in the compared models above is set to 256. We calculate the cumulated number of model parameters that are transmitted between the server and the clients in the first 20 communication rounds. Then, we early stop the running to report the average test accuracy across all clients at the 20th round. We show the results in Fig.~\ref{fig:trans_param}. In Fig.~\ref{fig:trans_param}, each point represents an architecture. We observe that FedMN has much less amount of transmitted parameters in the first 20 communication rounds, while its accuracy is higher than FedAvg. FedAvg does not perform well when the amount of the model parameters is large, i.e., 10-Layer MLPs at the 20th round because it convergences slowly when the model is complex. However, for each client, FedMN selects only partial model blocks. This decreases the model complexity, and thus converges faster than the model in FedAvg.

\section{Conclusion}
In this paper, we address the problem of joint distribution heterogeneity in the personalized FL. To solve it, we propose a novel FedMN approach that adaptively assembles architectures for each client by selecting a subset of module blocks from a module pool in the global model. The proposed FedMN adopts a light-weighted routing hypernetwork to model the joint distributions for each client and produce the module selection decisions. Advised by the decision, each client selects its personalized architecture. When federated updating, each client uploads, and downloads only part of the module parameters, which reduces the communication burden between the server and the clients. Extensive experiments show the effectiveness of the proposed approach compared with the state-of-the-arts. Analysis of parameters transmission shows that FedMN is a communication efficient method.

\section*{Acknowledgement}
This project was partially supported by NSF projects IIS-1707548 and
CBET-1638320.
\bibliographystyle{IEEEtran}
\begin{tiny}
\bibliography{IEEEtran}
\end{tiny}
\end{document}